# DeclareAligner: A Leap Towards Efficient Optimal Alignments for Declarative Process Model Conformance Checking

Jacobo Casas-Ramos[a,*], Manuel Lama[a], Manuel Mucientes[a]

[a]*Centro Singular de Investigación en Tecnoloxías Intelixentes (CiTIUS), Universidade de Santiago de Compostela, Santiago de Compostela, Spain*

**Abstract**

In many engineering applications, processes must be followed precisely, making conformance checking between event logs and declarative process models crucial for ensuring adherence to desired behaviors. This is a critical area where Artificial Intelligence (AI) plays a pivotal role in driving effective process improvement. However, computing optimal alignments poses significant computational challenges due to the vast search space inherent in these models. Consequently, existing approaches often struggle with scalability and efficiency, limiting their applicability in real-world settings. This paper introduces DeclareAligner, a novel algorithm that uses the A* search algorithm, an established AI pathfinding technique, to tackle the problem from a fresh perspective leveraging the flexibility of declarative models. Key features of DeclareAligner include only performing actions that actively contribute to fixing constraint violations, utilizing a tailored heuristic to navigate towards optimal solutions, and employing early pruning to eliminate unproductive branches, while also streamlining the process through preprocessing and consolidating multiple fixes into unified actions. The proposed method is evaluated using 8,054 synthetic and real-life alignment problems, demonstrating its ability to efficiently compute optimal alignments by significantly outperforming the current state of the art. By enabling process analysts to more effectively identify and understand conformance issues, DeclareAligner has the potential to drive meaningful process improvement and management.

## 1. Introduction

Process mining has become a crucial tool for organizations to analyze and improve their business processes, leveraging the increasing availability of event log data. Process mining encompasses a range of subjects (Aalst et al., 2012), including process discovery (reconstructing process models from event logs), conformance

*Corresponding author
*Email addresses:* jacobocasas.ramos@usc.es (Jacobo Casas-Ramos), manuel.lama@usc.es (Manuel Lama), manuel.mucientes@usc.es (Manuel Mucientes)

checking (comparing actual behavior with modeled behavior), and enhancement (improving process models based on insights gained).

Among these, conformance checking (Carmona et al., 2018) is particularly important as it compares an event log with a process model to identify deviations and discrepancies. One effective approach to conformance checking is through the use of alignments, which provide a detailed comparison of the executed steps in an event log with the expected behavior of the process model. Alignments visually represent the differences between the logged and modeled traces, highlighting discrepancies and mismatches. Optimal alignments (Adriansyah, 2014) further refine this approach by assigning a cost to each discrepancy and finding the alignment that minimizes the total cost.

Processes are often modeled using imperative approaches (Aalst, 1997), which spell out every possible allowed execution path in exhaustive detail. However, many real-world processes exhibit inherent variability, flexibility, and complexity, making it challenging to capture them using such rigid methods. In such cases, attempting to create an exhaustive model would likely result in a convoluted representation, characterized by numerous internal states —often referred to as a "spaghetti model" due to its intricate and cumbersome nature.

This is where declarative processes come into play (Back et al., 2018). Unlike their imperative counterparts, declarative process models focus on specifying the constraints that govern the behavior of a process, rather than dictating its exact execution flow. In essence, declarative models define "what" constraints must be satisfied during the execution of a process, without prescribing exactly "how" it should be done. This approach allows for greater flexibility in the actual process behavior, making it particularly suitable for capturing dynamic and complex processes. However, this flexibility also introduces challenges when it comes to conformance checking. As modeled constraints only restrict some relationships among activities, all unrestricted behavior is allowed by default. This means that a declarative model implicitly permits a multitude of different execution paths. The DECLARE language (Pesic et al., 2007) is widely regarded as one of the leading approaches to declarative process modeling, offering a flexible and concise way to define constraints on business processes.

While much research has been devoted to computing optimal alignments for imperative process models (Casas-Ramos et al., 2024; Dongen et al., 2017; Dongen, 2018; Lee et al., 2018; Sani et al., 2020; Taymouri and Carmona, 2020), there is a noticeable gap in optimizing alignment techniques for declarative models.

Simple declarative conformance checking approaches do not compute optimal alignments. These methods, such as (Chiariello et al., 2022) which is based on satisfiability problems, only report constraint failures without offering guidance on identifying the underlying issues. Others provide more detailed information, such as activations, fulfillments, and violations for each constraint (Burattin et al., 2016; Donadello et al., 2022; Maggi et al., 2019, 2011; Montali et al., 2013), but still fall short of computing optimal alignments. Specifically, they are unable to identify the root cause of non-conformances or provide a minimal way to fix them.

More advanced diagnostics computing optimal alignments have been published using finite-state automata (Giacomo et al., 2017; Leoni et al., 2012), but these methods can be computationally expensive due to the need to explore a large state space.

Despite these advancements, there remains a significant need for more efficient and scalable approaches to computing optimal alignments for declarative conformance checking. This paper presents DeclareAligner, a novel algorithm that builds upon the A* search algorithm (Hart et al., 1968). A* is a well-known method for finding the shortest path between two nodes in a graph. In the context of DeclareAligner, the nodes are referred to as states, and the graph is constructed incrementally by starting from an initial state representing the trace and iteratively applying actions to generate new states. These actions modify the previous state to make it increasingly compliant with the process model. By leveraging the flexibility of declare models, the approach is able to efficiently explore the vast search space of possible alignments to reach the optimal alignment. DeclareAligner offers several key innovations that enhance its performance and scalability:

- It only considers taking actions that directly contribute to repairing violated constraints.
- It proposes several search optimization strategies in order to manage difficult alignments:
  - An heuristic that merges the suggested actions for all violated constraints of a state to provide accurate cost estimates.
  - States that cannot reach the optimal alignment are pruned early to eliminate unfruitful branches of the search space.
  - Actions that are required by chain constraints and do not interfere with other constraints are applied before the search.
  - Actions perform multiple operations at once to avoid intermediate states.

The paper is structured as follows. Section 2 reviews existing approaches and highlights their limitations. Section 3 introduces the necessary concepts used by the algorithm. In Section 4 DeclareAligner is presented. Section 5 discusses the results of the empirical evaluation. Finally, Section 6 summarizes the contributions and suggests future work.

Table 1: Event log data collected for an e-learning process. Each row shows an event. An example trace with ID *Case234* is ⟨Enroll, Test, Exam⟩.

| Trace ID | Activity | Timestamp |
|---|---|---|
| Case234 | Enroll | 2023-09-01 22:16:29 |
| Case675 | Class | 2023-11-01 04:10:07 |
| Case234 | Test | 2023-11-15 02:09:48 |
| Case675 | Exam | 2023-12-13 07:24:41 |
| Case234 | Exam | 2024-01-19 06:42:33 |
| ⋮ | ⋮ | ⋮ |

## 2. Related work

Declarative conformance checking, particularly using the DECLARE language (Pesic et al., 2007), has gathered significant attention in recent years due to its ability to model complex and flexible business processes and reason about their behavior. Several approaches have been proposed for declarative conformance checking. These can be broadly categorized into two types: those that report only basic conformance information and those that compute optimal alignments to provide detailed insights into process deviations.

Methods in the first category, such as (Chiariello et al., 2022), simply execute recorded events in the process model and return a boolean result indicating conformance or non-conformance, without providing insights into specific issues. Other methods in this category, including (Burattin et al., 2016; Donadello et al., 2022; Maggi et al., 2019, 2011; Montali et al., 2013), report activations, satisfactions, and violations for each constraint within the DECLARE model and for every trace in the input log. While these results are faster to compute than optimal alignments, they do not identify root causes of non-conformance or suggest fixes. The main limitation of these approaches is that they only provide a high-level indication of conformance or non-conformance, without offering actionable insights for process improvement.

In contrast to simpler conformance checking methods, computing optimal alignments enables a more detailed analysis of process deviations and identification of opportunities for improvement. However, these approaches come at a higher computational cost. Currently, there are only two state-of-the-art approaches for computing optimal alignments of declarative models, namely those presented in (Leoni et al., 2012) and (Giacomo et al., 2017). A key characteristic shared by these approaches is the initial conversion of the DECLARE process model to finite-state automata. Specifically, standard DECLARE constraints can be translated into Linear-time Temporal Logic over finite traces ($LTL_f$) specifications, which in turn can be converted into finite-state automata for each constraint (Giannakopoulou and Havelund, 2001; Westergaard, 2011). The two state-of-the-art approaches differ primarily in their techniques for searching for the optimal alignment using these automata:

- The first approach by de Leoni et al. (Leoni et al., 2012) employs the A* algorithm to find optimal alignments, providing metrics such as fitness, precision, and generalization. However, this method can be computationally expensive due to the need to explore a large state space using a relatively simple heuristic.
- The second approach by de Giacomo et al. (Giacomo et al., 2017) converts automata to a planning problem and uses classical planners to find optimal alignments. This approach has shown better efficiency but may face scalability issues due to the required conversion to a planning task or the lack of control over the approach that the planner takes.

Existing methods for computing optimal alignments using DECLARE models struggle with large process models and logs. The proposed DECLAREALIGNER algorithm addresses these limitations by introducing a novel approach that focuses only on actions that can potentially resolve violated constraints, preprocesses the problem to reduce unnecessary work, and groups multiple fixes together into single actions to minimize redundant effort. Additionally, the algorithm utilizes an advanced heuristic and detects and removes dead-ends from the search space. This results in improved performance and scalability, enabling the tackling of more complex optimal alignment tasks.

## 3. Preliminaries

Event logs record sequences of events that happened during the execution of a process. A single occurrence of an activity is called an Event. Each event is characterized by a trace identifier, activity name, timestamp, and optional additional information. Events are grouped into traces, i.e., sequences of events that occur in a specific context or scope. In essence, a trace provides a snapshot of how a particular process instance has unfolded over time. For the purposes of this paper, only the ordered sequence of activities contained within each trace is necessary. An example log is shown in Table 1.

**Definition 1** (Trace). A trace is an ordered sequence of activities $\sigma = \langle A_1, \ldots, A_n \rangle$, where each activity $A_i$ is extracted from events that belong to the same case, i.e., that share the same trace identifier. They are sorted according to the timestamp of the associated event.

**Definition 2** (Log). An event log $L = [\sigma_1, \ldots, \sigma_n]$ is a multiset of traces.

The most widely used declarative process modeling language is DECLARE (Pesic et al., 2007). It defines a list of parametrized templates that, when instantiated with specific activities, become constraints restricting the execution of those activities. Table 2 provides a comprehensive overview of all supported DECLARE templates. For formal definitions, refer to (De Smedt et al., 2015). A DECLARE process model is simply a set of constraints.

Table 2: Supported DECLARE constraints and their natural language descriptions.

| | |
|---|---|
| *Existence(n, A)*: A occurs at least n times. | *Participation(A)*: A occurs at least once. |
| *Absence(n, A)*: A occurs at most n-1 times. | *AtMostOne(A)*: A occurs at most once. |
| *Exactly(n, A)*: A occurs exactly n times. | *Init(A)*: A is the first activity. |
| *End(A)*: A is the last activity. | *Choice(A, B)*: Either A or B, or both, occur. |
| *ExclusiveChoice(A, B)*: Either A or B, but not both, occur. | *RespondedExistence(A, B)*: If A occurs, B also occurs. |
| *Response(A, B)*: If A occurs, B follows. | *Precedence(A, B)*: If B occurs, A precedes it. |
| *AlternateResponse(A, B)*: If A occurs, B follows without any other A in between. | *AlternatePrecedence(A, B)*: If B occurs, A precedes it without any other B in between. |
| *ChainResponse(A, B)*: If A occurs, B is the next activity. | *ChainPrecedence(A, B)*: If B occurs, A is the previous activity. |
| *CoExistence(A, B)*: If A occurs, B also occurs, and vice versa. | *Succession(A, B)*: Combines Response and Precedence. |
| *AlternateSuccession(A, B)*: Combines AlternateResponse and AlternatePrecedence. | *ChainSuccession(A, B)*: Combines ChainResponse and ChainPrecedence. |
| *NotRespondedExistence(A, B)*: If A occurs, B does not. | *NotCoExistence(A, B)*: If A occurs, B does not, and vice versa. |
| *NotResponse(A, B)*: If A occurs, B does follow. | *NotPrecedence(A, B)*: If B occurs, A does not precede it. |
| *NotSuccession(A, B)*: If A occurs, B does not follow, and if B occurs, A does not precede it. | *NotChainResponse(A, B)*: A is not immediately followed by B. |
| *NotChainPrecedence(A, B)*: B is not immediately preceded by A. | *NotChainSuccession(A, B)*: A is not immediately followed by B and B is not immediately preceded by A. |

A constraint is considered violated when the specified condition or rule is not met within the trace. This means that the activities in the process occur in a manner that breaches the constraint. If a constraint is not violated, it is considered satisfied.

A few constraints such as Init(A) are always active. However, most constraints have an activation activity. If the activation does not appear in the trace, the constraint is always (vacuously) satisfied. For example, the activity A of Response(A, B) is the activation, as this constraint ensures that B must appear at some point after A, only if A appears in the trace.

Branching is an important part of the DECLARE language as it allows defining more complex constraints by inserting multiple activities instead of just one in each parameter of the template. When multiple branched

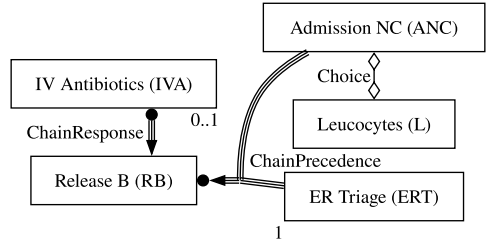

(a) Graphical representation.

Choice(ANC, L)
ChainPrecedence([ERT, ANC], RB)
Absence(2, IVA)
Exactly(1, ERT)
ChainResponse(IVA, RB)

(b) Text version.

Figure 1: Example of a DECLARE process.

activities are applied to a parameter of a template, any of them can trigger the effect associated with the parameter. Branched activities are shown between square brackets to avoid confusion.

**Example.** Figure 1 shows an example model from a hospital in both graphical and textual form with the following constraints:

- `Exactly(1, ERT)` specifies that `ERT` must occur exactly once in each trace.
- `Absence(2, IVA)` indicates that `IVA` cannot occur more than once.
- `Choice(ANC, L)` defines that either `ANC` or `L` must appear at some point in the trace.
- `ChainResponse(IVA, RB)` enforces that whenever `IVA` occurs, `RB` must also occur immediately after it.
- `ChainPrecedence([ANC, ERT], RB)` is an example of branching that specifies that whenever `RB` occurs, `ANC` or `ERT` must occur immediately before it.

An alignment maps a trace to a process model, highlighting conformance issues and enabling applications like model repair and auditing. An alignment is a sequence of legal moves that traverse both the trace and the process model from start to finish.

**Definition 3** (Legal move). Let $M$ be a DECLARE model with activities $A_M$, $L$ be a log with activities $A_L$, and $i$ be the first index (zero-based) of the trace that still needs to be aligned such that $\sigma[i]$ is the next activity to align. A move is a tuple $(a_L, a_M)$, where $a_L \in A_L \cup \gg$ and $a_M \in A_M \cup \gg$. Legal moves are moves of the following forms:

- **Synchronous moves** $(\sigma[i], \sigma[i])$ are available if the execution of the activity $a_n$ does not permanently violate any constraint of $M$. This updates the state of the model accordingly and advances the index $i$

Table 3: Alignment of the trace ⟨ANC, L, IVA, RB⟩ with the process from Figure 1. Activities are shown as their initials.

| LOG   | ANC | L | ≫   | IVA | RB |
|-------|-----|---|-----|-----|----|
| MODEL | ANC | L | ERT | ≫   | RB |

to the next trace activity.

- **Log moves** $(\sigma[i], \gg)$ are available if $i < |\sigma|$. This move increments $i$ by one, progressing in the trace without affecting the model.
- **Model moves** $(\gg, a)$ where $a \in A_M$ are available if all constraints of $M$ would not become permanently violated if $a$ is executed. This updates the state of the model by executing $a$.

Log and model moves may be referred to as asynchronous moves.

**Definition 4** (Alignment). An alignment $\gamma = \langle \gamma_1, \ldots, \gamma_n \rangle$ is a sequence of legal moves which reaches the end of the trace and a state of the model which satisfies all constraints.

Table 3 illustrates an example alignment, with each legal move as a column and the top and bottom rows represent the log and model parts respectively. A cost function assigns a penalty or cost to each move in an alignment, reflecting the degree of mismatch between the observed and modeled behavior. The standard cost function assigns a cost of 1 to asynchronous moves and a cost of 0 to synchronous ones. Table 3 has a total cost of 2.

**Definition 5** (Alignment cost, optimal alignment). A cost function $C : (A_L \cup \{\gg\}) \times (A_M \cup \{\gg\}) \setminus (A_L \times A_M \cup \{\gg\} \times \{\gg\}) \to (0, \infty)$ assigns a cost $c_{\gamma_i}$ to each possible legal asynchronous move, where $A_L$ is the set of activities in the trace and $A_M$ is the set of activities in the DECLARE model. Each valid alignment is assigned a cost $c_\gamma = c_{\gamma_1} + \cdots + c_{\gamma_n}$. An optimal alignment for a given trace and model is any of the alignments with the minimum total cost.

## 4. DeclareAligner algorithm

DECLAREALIGNER is built on top of A* search, a popular pathfinding and graph traversal algorithm that efficiently finds the shortest path between two nodes in a weighted graph. The A* search algorithm, known for its completeness and optimality guarantees, is particularly well-suited for the optimal alignments task as this task can be expressed as a graph that repairs violated constraints until all constraints are satisfied. The nodes of this graph are referred to as states.

The DECLAREALIGNER algorithm consists of several key components, as illustrated in Figure 2. The process commences with an initial state, which represents the current trace (Section 4.1). Prior to initiating the search for the optimal alignment, the algorithm performs preprocessing on the initial state (Section 4.5.2) effectively reducing the workload by starting closer to the solution. Subsequently, the A* search algorithm recursively explores a search space that is tailored to the computation of optimal alignments. The graph is incrementally constructed by applying fixes to violated constraints, thereby generating neighboring states (Section 4.2). This exploration is facilitated by various optimizations, including a tailored heuristic (Section 4.3), early pruning of dead-ends (Section 4.5.1), and the grouping of multiple fixes into single actions (Section 4.5.3). When a state that has no violated constraints is explored, the optimal alignment can be extracted (Section 4.4).

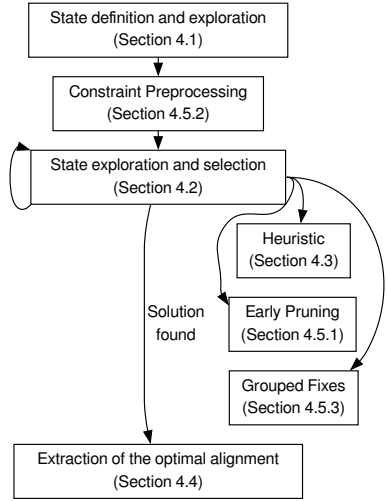

Figure 2: High-level overview of DeclareAligner.

*4.1. State definition and notation*

The algorithm explores a search space where each state is created by applying fixes to resolve constraint violations. A state contains the following attributes:

- **Cost** is a measure of the severity of fixes required to reach the state from the initial state.
- **Heuristic** is an estimate of the additional cost needed to transition from the current state to a goal state where all constraints are satisfied.
- **LTGraph** is a directed acyclic graph showing dependencies between activities or activity groups, with nodes connected by arcs that indicate precedence relationships. Activity groups can represent two types of relationships: branched activities, denoted as [A, B], where all activities have the same effect so they are interchangeable; or chained activity groups, denoted as A > B, where B directly follows A.
- **Violated activations** is a list of constraint activations that remain violated.
- **Will fix** is the activation that will be addressed next (Section 4.2).

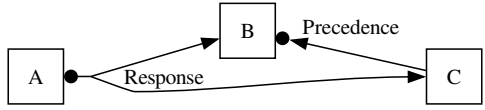

Figure 3: Process model of the running example.

The algorithm commences with an initial state, denoted as *state0*, which serves as the starting point for the search process. The initial state is constructed by creating an LTGraph that reflects the original problem instance: each activity of the trace is added as a node and connected to the node for the next activity of the trace to indicate their strict precedence. The initial state has a cost of 0 since no edits have been made yet.

**Example.** To facilitate understanding of the concepts presented, a running example will be utilized throughout this section. The running example consists of the trace $\langle A, A \rangle$ and the process model shown in Figure 3:

- `Response(A, [B, C])` enforces that if $A$ appears in the trace (activation), then either $B$ or $C$ must occur at any point after the activation.
- `Precedence(C, B)` specifies that if $B$ appears in the trace (activation), then $C$ must occur at any point before the activation.

The initial state for the running example is depicted in Figure 4, which shows from top to bottom:

- State identifier (*State0*), cost and heuristic values. The cost of the initial state is always 0, whereas the heuristic value is computed by identifying actions that are required to reach a goal state and adding their costs (Section 4.3).
- The LTGraph, which enforces the order between the two $A$ activities present in the trace. Distinct subscripts identify each activity instance to avoid ambiguity.
- Lists of violated activations for each process model constraint. The initial state reveals that $A_0$ and $A_1$ are violated activations of the response constraint, stemming from the absence of $B$ or $C$ activities succeeding each $A$ activity in the LTGraph. There are no violated activations for the precedence constraint, as there is no B in the LTGraph.
- The selected activation to repair, which in this case is $A_1$ (Section 4.2).

*4.2. State exploration and selection*

Neighbor generation is guided by identifying violated constraint activations and proposing repair actions. When multiple violations occur, those suggesting actions with the highest average costs are prioritized. This strategy of exploring high-cost actions first is grounded in the observation that non-zero cost actions result in asynchronous moves in the optimal alignment (Definition 3). Such actions can have a profound impact on other constraints, as they involve adding and removing activities. By prioritizing high-cost actions early in the search process, the risk of cascading effects that can arise when they are applied deeper in the search space, where numerous states are awaiting exploration, is mitigated. As a result of resolving high-cost actions first, subsequent repair actions suggested by other constraints tend to have localized effects on the state. This makes them more likely to be successfully repaired without triggering additional conflicts, and ultimately facilitating a faster convergence towards an optimal alignment.

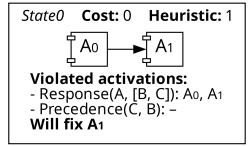

Figure 4: Initial state for the running example.

In cases where multiple violated activations have equal average costs, a secondary prioritization strategy to resolve ties is employed. Forward-looking constraints (e.g., Response) prioritize their last failed activation because all activations can be fixed if the target is inserted after this point. Analogously, backward-looking constraints favor their first failed activation. Negative constraints, however, require the opposite approach: since the goal is to avoid introducing targets altogether, removing the targets from the first activation effectively fixes all subsequent activations for forward-looking constraints and removing them from the last activation does so for backward-looking ones.

Generating neighboring states is accomplished by constructing actions, which comprise sequences of corrective modifications applied to the LTGraph in response to a violated constraint activation. All violations of declare constraints can be repaired by applying a combination of the following fix kinds:

- **Insert or remove node (Fig. 5a)**. Adds or deletes a node of the LTGraph, updating adjacent arcs when removing a node.
- **Add an arc (Fig. 5b)**. Enforces the origin node to precede the destination node.
- **Set start or end node (Fig. 5c)**. Sets a node as the first or last node, adding arcs to all other nodes and ensuring it remains in that position.
- **Subset of branched activities (Fig. 5d)**. In scenarios where a constraint is activated or violated by only a subset of branched activities present in a LTGraph node, this fix kind can be employed to avoid activation or enforce the target, respectively.
- **Merge or split chains (Fig. 5e)**. To efficiently implement chain constraints, the LTGraph nodes are capable of representing lists of possibly branched activities (separated by >). This fix kind merges two nodes, preserving their chained relationship without the need to add arcs to all other nodes.

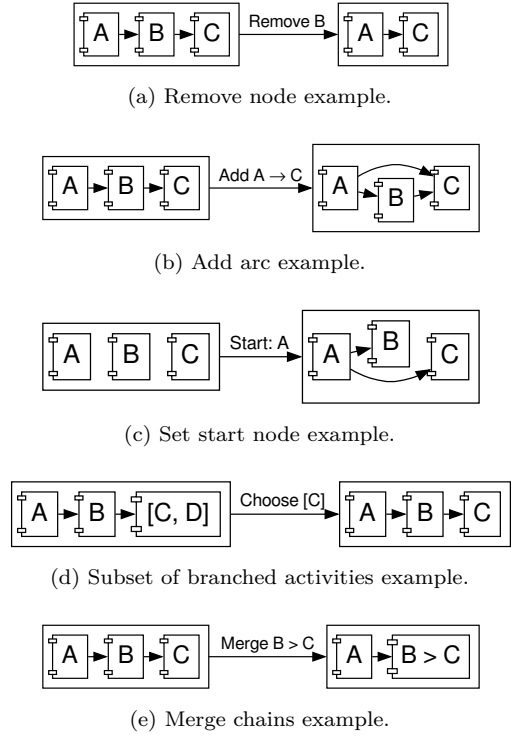

Figure 5: Examples of fix kinds applied to an LTGraph.

Only the insert or remove node fix kind has a non-null cost. This is because adding a node triggers a model move in the optimal alignment, while removing a node causes a log move (Section 4.4). In contrast, all other fix kinds only reduce the set of possible alignments that the LTGraph represents, without modifying their costs.

The standard cost function assigns a uniform cost of one to each insertion or removal operation. The cost of an action is the sum of the costs of its individual fixes. Each action taken contributes to increasing the cost of the current state.

**Definition 6** (State cost)**.** Let $T$ be the complete set of actions applied to the initial state in order to reach the current state $s$, and let $ct(t)$ be the cost of the action $t$, the cost of $s$ can be defined as:

$$cost(s) = \sum_{action \in T} ct(action)$$

The next state to explore is the previously unexplored state with the lowest total estimated cost ($cost + heuristic$). The algorithm continues to iteratively explore and select states until either a goal state is explored or no further states remain to be explored, in which case there is no optimal alignment.

**Example** (continued)**.** Figure 6 illustrates the complete search space explored by the algorithm to find the optimal alignment for the running example. The representation of an action is an arc connecting the parent state to the neighboring state. The arc is labelled with the fixes of the action. Goal states are highlighted with a green border, and the path to the optimal goal state is marked with green arrows.

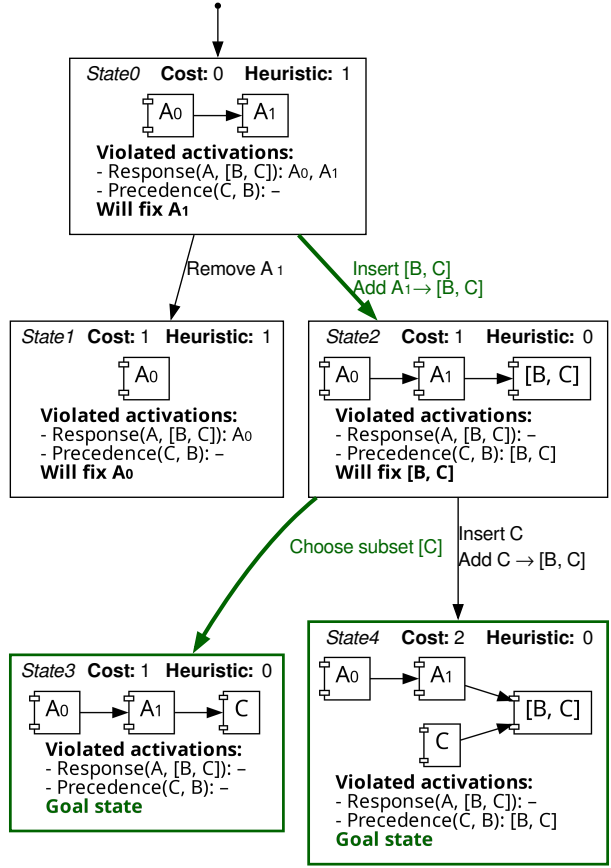

Figure 6: Search space discovered for the running example. Best viewed in color.

As previously mentioned the initial state (*state0*) has two violated activations for the response constraint. In this case, the algorithm selects the $A_1$ activation from the LTGRAPH to repair. There are only two possible fixes, which create the neighboring states of *state1* and *state2*:

- *State1* removes the activation so that the constraint is never activated, with a cost of 1.
- *State2* inserts the expected target of the constraint, which can be either of the activities $B$ or $C$. This results in the addition of the LTGRAPH node $[B, C]$ after the activation. This action results in satisfying both activations of the response constraint. However, it also activates the precedence constraint as one of the branched activities triggers it.

The algorithm selects *state2* to explore next based on its lower $cost + heuristic$ value. *state2* has only one violated activation to repair for the precedence constraint, resulting in the generation of the following neighboring states:

- *State3* removes the activation through a cost-free operation, which involves selecting a subset of branched activities to avoid triggering the precedence constraint.
- *State4* inserts the target activity before the activation. Note that the LTGraph now aggregates a series of possible activity orderings into a single state, reducing the size of the search space.

Ultimately, the algorithm reaches the goal state with the lowest cost, which in this case is *state3*. The optimal alignment can then be extracted from the goal state (Section 4.4).

*4.3. Heuristic*

The heuristic function used in DeclareAligner leverages knowledge about required repair actions to provide an accurate estimate of the remaining cost to reach a goal state. Given that all violated activations must be addressed, this approach is based on two key insights:

- **Minimum required actions**. At least one action from the set of repair actions suggested for each violated activation is necessary to resolve the violation.
- **Optimistic merging**. Actions that can fix multiple violations must be merged optimistically to avoid overestimating the remaining cost.

The heuristic function, formally defined in Algorithm Algorithm 1, takes advantage of these insights to compute an accurate estimate of the remaining cost. The process begins by identifying all violated activations in the current state (alg. 1:2). For each violated activation, the set of proposed repair actions is retrieved (alg. 1:3).

Next, Dijkstra's algorithm (Dijkstra, 1959) is used to search for the combination of one action from each set that incurs the lowest total cost (alg. 1:4). The main loop of the search (alg. 1:8) involves four key functions:

- The RemoveLeastCost function (alg. 1:9) deletes and returns the unexplored heuristic state with the lowest total cost from the set of *hstates*, where the cost is determined by the sum of action costs. This state becomes the next candidate for exploration in the search process.
- The HActions function (alg. 1:12) selects the next violated activation that was not previously explored and adds all of its suggested fixes as new actions to discover new neighboring states.
- The HExplore function (alg. 1:14) discovers a new *child* state from each of the proposed actions of the HActions function. It does so by adding the selected action to the current state and merging it with existing actions to avoid overestimating the cost.

**Algorithm 1** Heuristic function
---
1: **function** HEURISTIC(*state*)
2:     *activations* ← violatedActivations(*state*)
3:     *actionSets* ← {actions(*a*) | *a* ∈ *activations*}
4:     *combination* ← HDIJKSTRA(*actionSets*)
5:     **return** $\sum_{action \in combination} ct(action)$
6: **function** HDIJKSTRA(*actionSets*)
7:     *hstates* ← {{}}     ▷ Set of unexplored *hstates*
8:     **loop**     ▷ Main loop of Dijkstra's algorithm
9:         *hstate* ← REMOVELEASTCOST(*hstates*)
10:         **if** *hstate* is a goal state **then**
11:             **return** *hstate*
12:         *hactions* ← HACTIONS(*hstate*, *actionSets*)
13:         **for all** *haction* ∈ *hactions* **do**
14:             *child* ← HEXPLORE(*hstate*, *haction*)
15:             ADDORREPLACE(*hstates*, *child*)

- The ADDORREPLACE function (Alg. 1:15) checks whether the generated *child* state is already in *hstates*. If not, it adds the *child*; otherwise, it replaces the existing *hstate* only if the new one has a lower cost.

The search algorithm concludes upon exploring the first *hstate* that meets the goal criteria, namely, when the *hstate* includes at least one action from each activation (alg. 1:10). This is always found since any combination that includes actions from all activations is valid, and Dijkstra's algorithm ensures that the optimal one is found (alg. 1:11). Hence, the returned *hstate* is the combination of actions from all violated activations that yields the lowest cost. The resulting heuristic value is determined by this minimum possible cost of the combined actions (alg. 1:5). This value provides an informed estimate of the effort required to resolve all outstanding issues and achieve a valid solution.

**Example** (continued). Consider *state0* from Figure 6. Two similar violated activations propose actions that either remove the activation or insert a new node after it, both of cost 1. However, the insert actions can be combined to solve both violations simultaneously. The heuristic searches for the lowest-cost combination of actions, which in this case is a single insert action of cost 1. This yields a heuristic value of 1 for *state0*.

*4.4. Extraction of the optimal alignment*

To extract the optimal alignment from the goal state, a topological sort for the LTGRAPH is found. A topological sort is a linear ordering of nodes in the LTGRAPH such that for every edge, the starting node comes before the ending node in the order. This results in a sequence of activities executed in the model, where branched activities are resolved by selecting any one of them.

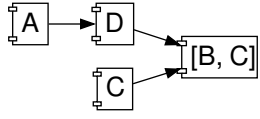

Figure 7: Example of an LTGraph with branched activities.

Table 4: Example of optimal alignment extracted from the LTGraph shown in Figure 7 and the trace $\langle A_0, D, A_1 \rangle$ .

| LOG | A | ≫ | D | ≫ | A |
|---|---|---|---|---|---|
| MODEL | A | C | D | C | ≫ |

We then process both the nodes from the topological sort and the original trace simultaneously to extract the optimal alignment:

- If an LTGraph node was inserted when performing an action, a model move is added to the alignment and advance the topological sort.
- Otherwise:
    - If the next LTGraph node matches the next activity in the trace, a synchronous move is added to the alignment and advance both.
    - Otherwise, a log move is added to the alignment and advance the trace.

Any remaining activities in the trace are handled by adding them as log moves. The collected sequence of moves is the optimal alignment.

**Example.** Let $\langle A, D, A \rangle$ be the trace and Figure 7 be the LTGraph of the goal state from which the optimal alignment will be extracted. A valid topological sort of the LTGraph is $\langle A_0, C, D, C \rangle$, after selecting $C$ from the branched activities $[B, C]$. This yields an optimal alignment where $A_0$ is a synchronous move, followed by a model move for the inserted activity $C$, another synchronous move for $D$, and another model move for the inserted activity $C$. Finally, since the last $A$ from the trace was not processed, it results in a log move, producing the optimal alignment shown in Table 4.

*4.5. Optimizations*

Several optimization techniques can further improve the efficiency and effectiveness of the DeclareAligner algorithm. These enhancements build upon the foundation established in the previous sections and are designed to refine the performance of the algorithm.

*4.5.1. Early Pruning*

The DeclareAligner algorithm improves search efficiency by pruning branches that cannot lead to a goal state. If any violated activation has no repair action, it is impossible to reach a goal state from that point. The reason behind this is that the violated activation will never be repaired even if all other violated activations are eventually repaired through the exploration of the search space. Since the heuristic already computes the suggested actions for all violated activations, the detection of dead-ends incurs no additional overhead.

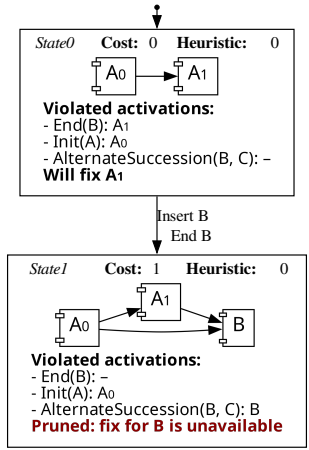

Figure 8: Illustrative example highlighting the advantages of enabling the Early Pruning optimization, cutting down the search space from 7 discovered states to just 2. Best viewed in color.

To determine whether an action is feasible for a given state, a cycle detection algorithm is executed on the resulting LTGRAPH after applying the action. This is because the LTGRAPH represents a strict order, and any cycles would result in an impossible topological sort. In other words, if a cycle were present, it would indicate that there are conflicting requirements in the graph, making it impossible to extract a valid alignment from the given state. By detecting such cycles, actions that would lead to inconsistencies can be identified and pruned.

**Example.** The example illustrated in Figure 8 demonstrates the benefits of early pruning. Initially, the end constraint is repaired by inserting a $B$ activity at the end of the LTGRAPH. Upon exploring this second state, it is identified that the alternate succession constraint would require the insertion of a $C$ after the $B$, and this cannot be satisfied without creating a cycle in the LTGRAPH ($B \rightleftarrows C$), indicating conflicting requirements. As a result, this state is recognized as a dead-end, since repairing the alternate succession constraint is necessary to reach a goal state.

Without early pruning, the algorithm would proceed by addressing the violated $A_0$ activation of the init constraint, leading to the discovery of numerous additional neighbors before ultimately realizing that each of these states cannot reach a goal state due to the impossibility of solving the alternate succession constraint. In this simple scenario, enabling early pruning reduces the search space from 7 states to 2 states. For problems with more constraints, the difference is even more pronounced, as checking all constraints for dead-ends

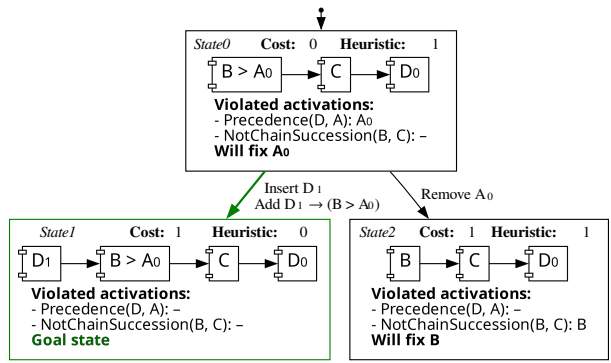

Figure 9: This example shows how Constraint Preprocessing simplifies the search space by decreasing the number of discovered states from 9 to just 3. Best viewed in color.

prevents unnecessary exploration and avoids realizing later that no solution exists.

*4.5.2. Constraint Preprocessing*

The DeclareAligner algorithm can leverage preprocessing techniques to reduce problem complexity for certain constraints, specifically those containing "chain" in their name. By merging nodes in the LTGraph of the initial state that are affected by chain constraints, the algorithm can significantly improve efficiency while maintaining optimality. To achieve this, care must be taken to ensure that merged nodes can be split later if necessary.

This preprocessing step involves combining two consecutive chained activities into a single node in the initial state if they satisfy the constraint. By applying this preprocessing step before initiating the search, a substantial number of states can be eliminated from consideration, resulting in improved overall efficiency.

**Example.** Figure 9 illustrates the impact of this optimization on the search space. In this example, the process model requires that immediately after an $B$, a $C$ does not appear. Since this condition is met in the initial trace, the optimization merges the nodes for $B$ and $A_0$ into a single LTGraph node.

If Constraint Preprocessing were disabled, the nodes of the LTGraph would still need to be merged to enforce the not chain succession constraint, necessitating an intermediate action. Consequently, the search space for this simple example would consist of 9 states instead of just 3 states. Notably, this optimization enables chain constraints to behave similarly to most other constraints: if there is no issue with the original trace, there is no need to generate repair actions during the search.

*4.5.3. Grouped Fixes*

This optimization technique involves consolidating multiple fixes into a single action, thereby eliminating the need to propose individual fixes separately. This approach is particularly effective when dealing with complex actions that require a specific sequence of fixes. For instance, in cases where an activity must appear a certain number of times in the trace, grouping the necessary fixes into a single action allows for efficient insertion of all the required occurrences simultaneously.

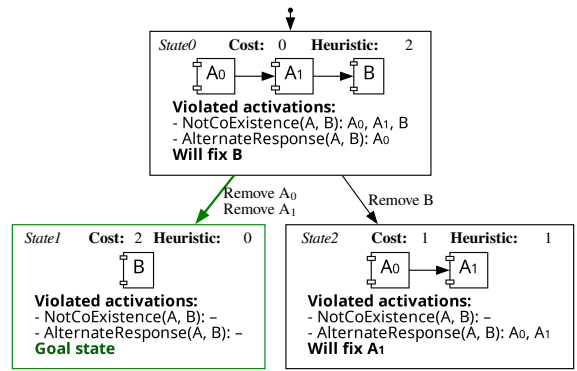

Figure 10: Example search space demonstrating the benefits of the Grouped Fixes optimization, which reduces the number of discovered states from 7 to 3. Best viewed in color.

Consolidating these fixes has a significant impact on the search space. By avoiding the generation of intermediate states, the algorithm reduces the need to explore many more unnecessary combinations of fixes.

**Example.** Figure 10 illustrates the benefits of this optimization technique. Consider a scenario where the `NotCoexistence(A, B)` constraint is violated because both $A$ and $B$ appear in the trace. In this case, the only possible actions to resolve the violation are removing all instances of $A$ or removing all instances of $B$. Thanks to grouped fixes, these actions can be performed in a single exploration step, allowing the algorithm to reach the goal state after discovering only 3 states.

In contrast, disabling this optimization would result in a significantly larger search space. For this simple example, the search space would consist of 7 states instead of just 3, highlighting the effectiveness of grouped fixes in improving the efficiency of the algorithm.

## 5. Evaluation

DeclareAligner is implemented in Kotlin and executed on the same Java Virtual Machine[1] as the other state-of-the-art algorithms that have been tested. Experiments are conducted on an Intel Xeon Gold 5220R CPU in single-threaded mode with 4GiB of RAM. The source code, dataset and binaries are available online[2].

---

[1] OpenJDK Temurin-22.0.1+8
[2] https://apps.citius.gal/ltgraph/

Table 5: Ablation study results. `Time` and `ExpStates` represent the average over the complete dataset. `Timeouts` is the number of trace-model pairs for which the time limit was reached. `Reduction` values indicate improvements relative to the baseline algorithm, expressed as percentages for `Time` and `ExpStates`, and absolute differences for `Timeouts`.

| EP | CP | GF | Time (s) | Time reduction | ExpStates | ExpStates reduction | Timeouts | Timeout reduction |
|---|---|---|---|---|---|---|---|---|
|  |  |  | 101.6 | — | 3,802.7 | — | 2,803 | — |
| ✓ |  |  | 17.3 | 83.0% | 432.5 | 88.6% | 358 | 2,445 |
|  | ✓ |  | 76.6 | 24.6% | 3,030.2 | 20.3% | 1,968 | 835 |
|  |  | ✓ | 53.8 | 47.0% | 1,210.3 | 68.2% | 1,164 | 1,639 |
| ✓ | ✓ |  | 55.2 | 45.6% | 792.4 | 79.2% | 1,352 | 1,451 |
| ✓ |  | ✓ | 36.0 | 64.5% | 186.1 | 95.1% | 698 | 2,105 |
|  | ✓ | ✓ | 17.3 | 83.0% | 470.8 | 87.6% | 357 | 2,446 |
| ✓ | ✓ | ✓ | **7.6** | **92.5%** | **109.7** | **97.1%** | **113** | **2,690** |

*5.1. Datasets*

The evaluation utilizes the same datasets as those in (De Giacomo et al., 2023), extended with additional alignment problems to include all DECLARE constraint templates.

- **Real-life dataset from (De Giacomo et al., 2023)**. A personal loan application process log from a Dutch financial institute serves as the real-life dataset. Its process model contains 16 constraints and the dataset has a total of 854 trace-model pairs.
- **Synthetic dataset from (De Giacomo et al., 2023)**. Synthetic logs are generated using three DECLARE models with 10, 15, and 20 constraints. Noise is introduced by replacing 3, 4, or 6 constraints with their negative counterparts. The log generator from (Di Ciccio et al., 2015) produces four logs for each modified model, containing 100 traces of varying lengths (1-50, 51-100, 101-150, and 151-200 events). These noisy logs are aligned with the original models, totaling 3,600 trace-model pairs.
- **Extended dataset**. 16 DECLARE templates were missing from the original test datasets, including alternate relation constraints, choice constraints, and various negative constraints. Additional synthetic log-model pairs are generated following the same procedure described in the previous paragraph, but ensuring all constraint templates are present in the results. This adds another 3,600 trace-model pairs.

The complete dataset contains 8,054 trace-model pairs.

*5.2. Ablation study*

This ablation study evaluates the impact of each component in DECLAREALIGNER to determine how they individually contribute to its overall performance. Various metrics are collected for each trace-model pair, such as execution time (`Time`), expanded states (`ExpStates`), and the timeouts (`Timeouts`). The results of the ablation study are shown in Table 5 and are discussed in the following subsections.

Early Pruning (Section 4.5.1) results in a large decrease of 83.0% in average execution times. Furthermore, it enables the computation of many more alignments, reducing the number of timeouts by 2,445.

This improvement is largely due to the pruning of a substantial number of states that would have led to unsolvable constraints, reducing the number of expanded states by 88.6%. By avoiding unnecessary computations and reducing the search space, this optimization ultimately is capable of computing complex alignments that would not be possible otherwise.

Constraint Preprocessing (Section 4.5.2) yields a significant decrease in execution time (24.6%). This improvement is accompanied by a considerable reduction in the number of timeouts, with 835 fewer cases timing out.

The primary reason for this improvement is that constraint preprocessing allows the algorithm to start from a more advanced initial state, thereby avoiding the generation of numerous unnecessary states (20.3%). By performing shortcuts for highly likely operations upfront, this optimization reduces the overall search space, making it possible to compute alignments more efficiently.

Grouped Fixes (Section 4.5.3) leads to a substantial reduction in execution time (47.0%) and a decrease in timeouts by 1,639 cases. This improvement is primarily due to the fact that grouped fixes enable the algorithm to prune the search space more effectively, reducing the number of expanded states by 68.2% and avoiding unnecessary exploration of intermediate states and their neighbors.

Pairing any two optimizations generally leads to positive outcomes. This can be attributed to the fact that each optimization targets a distinct aspect of the search space, thereby reducing the number of expanded states required. Constraint Preprocessing focuses on moving the initial state closer to the goal, whereas Grouped Fixes avoid intermediate states when expanding neighbors. Meanwhile, Early Pruning eliminates dead-ends in the search space, preventing unnecessary exploration.

Combining all three optimizations yields the most substantial improvements across all tested metrics. The average execution time decreases by 92.5%, the expanded states are reduced by 97.1%, and timeouts are reduced by 2,690 cases. By integrating these three optimizations, DeclareAligner is able to leverage their complementary strengths, resulting in a more efficient search strategy.

*5.3. State-of-the-art comparison*

Table 6: Comparison of the average execution times and number of timeouts reached for each tested algorithm on the evaluated dataset.

| Algorithm | Time (s) | Timeouts |
|---|---|---|
| PlannerFD | 111.3 | 1,947 |
| DeclareReplayer | 72.5 | 1,184 |
| PlannerBA | 46.1 | 400 |
| DeclareAligner | **7.6** | **113** |

The state-of-the-art algorithms capable of solving the alignment problem for DECLARE processes are DeclareReplayer (Leoni et al., 2012) and PlannerBA/PlannerFD (Giacomo et al., 2017). Table 6 summarizes the results of the state-of-the-art approaches and DeclareAligner on the described dataset.

The results demonstrate a significant improvement over the state-of-the-art techniques, with substantial reductions in execution time. Further analysis shows that most of the cases where the execution time of DeclareAligner is close to the state of the art are the simplest alignment problems in the dataset. This can be attributed to the initialization phase required by some proposed optimizations, which can introduce a delay for trivial alignments but prove highly beneficial for complex instances.

To further substantiate the comparison, a ranking of all algorithms based on their execution times is computed. For each sample, the algorithms are ranked, with ties being assigned the average rank. These ranks are then averaged across all samples to obtain an overall ranking for each algorithm.

To ensure that the results are statistically significant, the Friedman test is performed to assess overall differences between the algorithms, followed by the Holm post-hoc test to identify specific pairwise differences. The ranking and statistical tests are presented in Table 7.

As expected, the ranks show the clear advantage in execution times of DeclareAligner. The p-values indicate that the differences in performance among the algorithms are statistically significant, demonstrating that the proposed approach outperforms others in terms of computational efficiency.

To illustrate the efficiency and scalability of DeclareAligner compared to state-of-the-art algorithms, Figure 11 shows the number of trace-model pairs (y-axis) that can be solved within a given execution time (x-axis). Each algorithm is depicted as a distinct line, allowing for easy comparison of their performance across different time thresholds and number of logs solved.

Figure 11 reveals that DeclareAligner significantly outperforms the state-of-the-art algorithms in terms

Table 7: Ranking and statisical tests comparing execution times over the complete dataset.

(a) Average ranking of each tested algorithm.

| Algorithm | Rank |
|---|---|
| DeclareAligner | **1.15** |
| DeclareReplayer | 2.58 |
| PlannerBA | 2.97 |
| PlannerFD | 3.30 |

(b) Results of the Friedman test and Holm post-hoc analysis, indicating overall differences between algorithms and specific pairwise differences.

| Statistical test | p-value |
|---|---|
| DeclareAligner vs all | $< 10^{-6}$ |
| DeclareAligner vs DeclareReplayer | $< 10^{-6}$ |
| DeclareAligner vs PlannerBA | $< 10^{-6}$ |
| DeclareAligner vs PlannerFD | $< 10^{-6}$ |

of execution time. DeclareReplayer is also really fast to align the simplest trace-model pairs, even being capable of outperforming DeclareAligner for some of the most straightforward alignments. However, its performance quickly degrades with complexity of the input problem, showing the superior scalability of DeclareAligner. DeclareReplayer loses the position of 2nd in number of trace-model pairs solved at around 17s, where PlannerBA overtakes it. This happens because both planner-based techniques take a little while to convert the alignment task into a classic planning problem and the planner also needs to perform some preprocessing optimizations in order to solve harder problems faster.

A notable achievement of DeclareAligner is its ability to align 90% of the tested trace-model pairs in 3.5 seconds or less per alignment, outperforming the next best state-of-the-art algorithm (PlannerBA) which requires up to 100 seconds per problem to reach the same number of solutions. Furthermore, the proposed optimizations enable DeclareAligner to successfully compute alignments for 231 log-model pairs that remain unsolved by all other algorithms within a 5-minute time limit.

## 6. Conclusions

This work introduces DeclareAligner, an A*-based algorithm for efficiently computing optimal alignments of declarative process models. The contributions of this research are twofold. Firstly, a novel algorithmic approach has been proposed to leverage the inherent flexibility of declarative processes in calculating optimal alignments. Secondly, additional search optimizations and techniques have been developed and integrated to effectively handle complex scenarios.

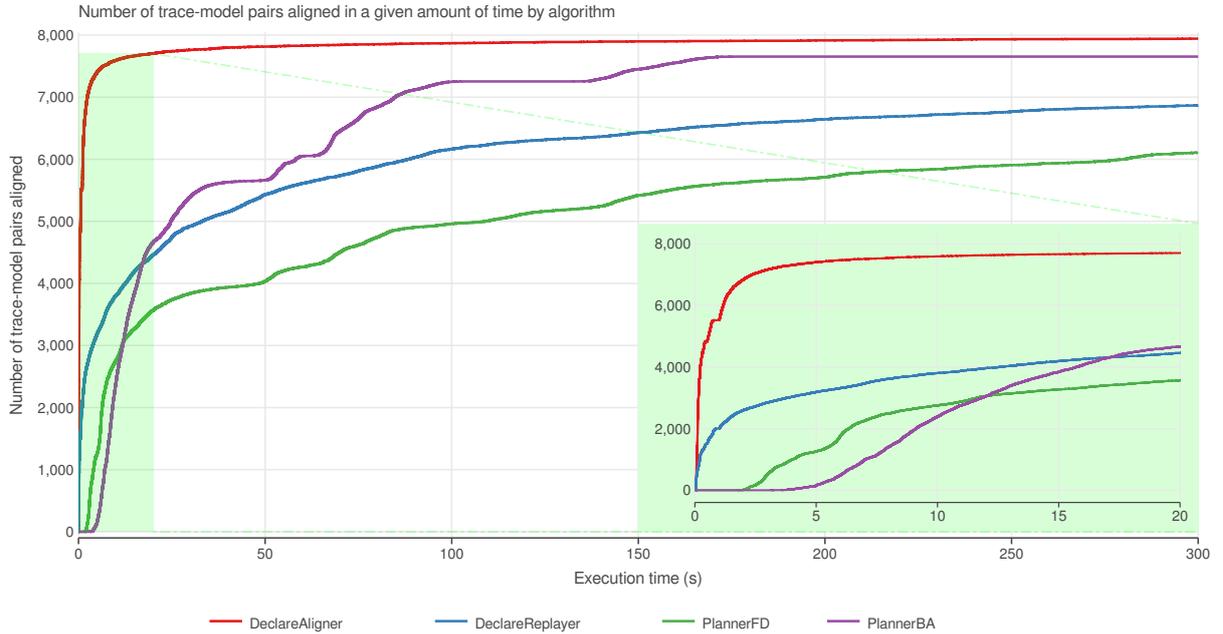

Figure 11: Number of trace-model pairs successfully aligned within a given time frame by each algorithm. The zoomed-in section in the bottom-right corner highlights the differences for shorter time intervals. Best viewed in color.

The evaluation, conducted on 8,054 trace-model pairs from real-world and synthetic processes, demonstrates the substantial performance improvements of the proposed method over the state of the art. Notably, DeclareAligner successfully aligns 7,943 pairs within the 5 minutes time limit, outperforming PlannerBA (7,654), DeclareReplayer (6,870), and PlannerFD (6,107). Moreover, the proposed optimizations enable DeclareAligner to compute alignments for 231 pairs that are unsolvable by other algorithms within the same time frame. The results show that DeclareAligner achieves faster alignment times than any other algorithm for 7,014 of the evaluated trace-model pairs. The significance of this research lies in its ability to facilitate efficient conformance checking while preserving the accuracy of optimal alignments, ultimately leading to improved process quality and reduced costs through effective detection and diagnosis of log-related issues.

In future research, the focus will shift towards adapting the proposed algorithm to accommodate data-aware declarative process models, ensuring retention of its high performance. Through extension to manage intricate data attributes and relationships, the goal is to offer a scalable conformance checking solution suitable for real-world applications demanding both precision and efficiency.

## 7. CRediT authorship contribution statement

**Jacobo Casas-Ramos**: Conceptualization, Methodology, Software, Validation, Formal analysis, Investigation, Data curation, Writing - Original Draft, Writing - Review & Editing, Visualization. **Manuel Lama**: Conceptualization, Methodology, Resources, Writing - Review & Editing, Supervision, Funding acquisition. **Manuel Mucientes**: Conceptualization, Methodology, Resources, Writing - Review & Editing, Supervision, Funding acquisition.

## 8. Acknowledgement

This research was partially funded by the Spanish Ministerio de Ciencia e Innovación [grant numbers PID2023-149549NB-I00, TED2021-130374B-C21]. These grant are co-funded by the European Regional Development Fund (ERDF). Jacobo Casas-Ramos is supported by the Spanish Ministerio de Universidades under the FPU national plan [grant number FPU19/06668].

## 9. References

Aalst, W. van der, Adriansyah, A., Medeiros, A.K.A. de, Arcieri, F., Baier, T., Blickle, T., Bose, J.C., Brand, P. van den, Brandtjen, R., Buijs, J., others, 2012. Process mining manifesto. In: Business Process Management Workshops: BPM 2011 International Workshops. Springer, pp. 169–194.

Aalst, W.M.P. van der, 1997. Verification of workflow nets. In: Azéma, P., Balbo, G. (Eds.), Application and Theory of Petri Nets 1997. Springer Berlin Heidelberg, Berlin, Heidelberg, pp. 407–426.

Adriansyah, A., 2014. Aligning observed and modeled behavior.

Back, C.O., Debois, S., Slaats, T., 2018. Towards an empirical evaluation of imperative and declarative process mining. In: Woo, C., Lu, J., Li, Z., Ling, T.W., Li, G., Lee, M.L. (Eds.), Advances in Conceptual Modeling. Springer International Publishing, Cham, pp. 191–198.

Burattin, A., Maggi, F.M., Sperduti, A., 2016. Conformance checking based on multi-perspective declarative process models. Expert systems with applications 65, 194–211.

Carmona, J., van Dongen, B., Solti, A., Weidlich, M., 2018. Conformance checking: Relating processes and models. Springer.

Casas-Ramos, J., Mucientes, M., Lama, M., 2024. REACH: Researching efficient alignment-based conformance checking. Expert Systems with Applications 241, 122467.

Chiariello, F., Maggi, F.M., Patrizi, F., 2022. ASP-based declarative process mining. In: Thirty-Sixth AAAI Conference on Artificial Intelligence (AAAI 2022). AAAI Press, pp. 5539–5547.

De Giacomo, G., Fuggitti, F., Maggi, F.M., Marrella, A., Patrizi, F., 2023. A tool for declarative trace alignment via automated planning. Software Impacts 16, 100505.

De Smedt, J., Vanden Broucke, S.K.L.M., De Weerdt, J., Vanthienen, J., 2015. A full r/i-net construct lexicon for declare constraints. SSRN Electronic Journal.

Di Ciccio, C., Bernardi, M.L., Cimitile, M., Maggi, F.M., 2015. Generating event logs through the simulation of declare models. In: Barjis, J., Pergl, R., Babkin, E. (Eds.), Enterprise and Organizational Modeling and Simulation. Springer International Publishing, Cham, pp. 20–36.

Dijkstra, E.W., 1959. A note on two problems in connexion with graphs. Numerische mathematik 1, 269–271.

Donadello, I., Riva, F., Maggi, F.M., Shikhizada, A., 2022. Declare4Py: A Python library for Declarative Process Mining. In: Proceedings of the Best Dissertation Award, Doctoral Consortium, and Demonstration & Resources Track BPM 2022, CEUR Workshop Proceedings. CEUR-WS.org, pp. 117–121.

Dongen, B. van, Carmona, J., Chatain, T., Taymouri, F., 2017. Aligning modeled and observed behavior: A compromise between computation complexity and quality. In: 29th International Conference on Advanced Information Systems Engineering (CAiSE). Springer, pp. 94–109.

Dongen, B.F. van, 2018. Efficiently computing alignments. In: Business Process Management. Springer International Publishing, Cham, pp. 197–214.

Giacomo, G.D., Maggi, F.M., Marrella, A., Patrizi, F., 2017. On the disruptive effectiveness of automated planning for LTLf-based trace alignment. Proceedings of the AAAI Conference on Artificial Intelligence 31.

Giannakopoulou, D., Havelund, K., 2001. Automata-based verification of temporal properties on running programs. In: Proceedings 16th Annual International Conference on Automated Software Engineering (ASE 2001). IEEE Comput. Soc.

Hart, P., Nilsson, N., Raphael, B., 1968. A formal basis for the heuristic determination of minimum cost paths. IEEE Transactions on Systems Science and Cybernetics 4, 100–107.

Lee, W.L.J., Verbeek, H.M.W., Munoz-Gama, J., Aalst, W.M.P. van der, Sepúlveda, M., 2018. Recomposing conformance: Closing the circle on decomposed alignment-based conformance checking in process mining. Information Sciences 466, 55–91.

Leoni, M. de, Maggi, F.M., Aalst, W.M.P. van der, 2012. Aligning event logs and declarative process models for conformance checking. In: Lecture Notes in Computer Science. Springer Berlin Heidelberg, pp. 82–97.

Maggi, F.M., Montali, M., Bhat, U., 2019. Compliance monitoring of multi-perspective declarative process models. In: Dijkman, R., Si-Said, S. (Eds.), Proceedings of the 23rd IEEE International Enterprise Distributed Object Computing Conference (EDOC 2019). IEEE Computer Society, pp. 151–160.

Maggi, F.M., Montali, M., Westergaard, M., Aalst, W.M.P. van der, 2011. Monitoring business constraints with linear temporal logic: An approach based on colored automata. In: International Conference on Business Process Management. Springer; Springer Berlin Heidelberg, pp. 132–147.

Montali, M., Maggi, F., Chesani, F., Mello, P., Aalst, W., 2013. Monitoring business constraints with the event calculus. ACM Transactions on Intelligent Systems and Technology 5, 1–30.

Pesic, M., Schonenberg, H., Aalst, W.M.P. van der, 2007. DECLARE: Full support for loosely-structured processes.

Sani, M.F., Zelst, S.J. van, Aalst, W.M.P. van der, 2020. Conformance checking approximation using subset selection and edit distance. In: 32nd International Conference on Advanced Information Systems Engineering (CAiSE). Springer, pp. 234–251.

Taymouri, F., Carmona, J., 2020. Computing alignments of well-formed process models using local search. ACM Transactions on Software Engineering and Methodology 29, 1–41.

Westergaard, M., 2011. Better algorithms for analyzing and enacting declarative workflow languages using LTL. In: Business Process Management: 9th International Conference, BPM 2011, Clermont-Ferrand, France, August 30-September 2, 2011. Proceedings 9. Springer, pp. 83–98.